\newcommand{\wl}[1]{#1}
\documentclass[journal]{IEEEtran}
\usepackage{amsmath}
\usepackage{amssymb}
\usepackage{color}
\usepackage{mathrsfs}

\usepackage{cite}
\usepackage[pdftex]{graphicx}

\ifCLASSINFOpdf
\else
\fi
\usepackage{algorithmic}
\begin{document}
\title{Dense Video Captioning using Graph-based Sentence Summarization}

\author{Zhiwang~Zhang,
        Dong~Xu,~\IEEEmembership{Fellow,~IEEE,}
        Wanli~Ouyang,~\IEEEmembership{Senior Member,~IEEE,}
        and Luping~Zhou,~\IEEEmembership{Senior Member,~IEEE}
        
\thanks{This research is supported by the Australian Research Council Future Fellowship under Grant FT180100116.(Corresponding author: Dong Xu.)}                
\thanks{Z. Zhang, D. Xu, W. Ouyang and L. Zhou are with the School of Electrical and Information Engineering, University of Sydney, Sydney, NSW, 2008 Australia e-mail: (zhiwang.zhang@sydney.edu.au, dong.xu@sydney.edu.au, wanli.ouyang@sydney.edu.au, and luping.zhou@sydney.edu.au ).}
% <-this % stops a space
}

%\thanks{Manuscript received April 19, 2005; revised December 27, 2012.}}


\markboth{IEEE TRANSACTIONS ON XXXX, ~Vol.XX, No.XX, XXXX,XXXX}%
{Shell \MakeLowercase{\textit{et al.}}: Bare Demo of IEEEtran.cls for Journals}



\maketitle
%\IEEEpubid{\begin{minipage}{\textwidth}\vspace{10mm}\ \\[8pt] \centering
 %1520-9210 (c) 2020 IEEE. Personal use is permitted, but republication/redistribution requires IEEE permission. See http://www.ieee.org/publications$\_$standards/publications/rights/index.html for more information.
%\end{minipage}} 
\begin{abstract}
Recently, dense video captioning has made attractive progress in detecting and captioning all events in a long untrimmed video. Despite promising results were achieved, most existing methods do not sufficiently explore the scene evolution within an event temporal proposal for captioning, and therefore perform less satisfactorily when the scenes and objects change over a relatively long proposal. To address this problem, we propose a graph-based partition-and-summarization (GPaS) framework for dense video captioning within two stages. For the ``partition" stage, a whole event proposal is split  into short video segments for captioning at a finer level. For the ``summarization" stage,  the generated sentences carrying rich description information for each segment are summarized into one sentence to describe the whole event. We particularly focus on the ``summarization" stage, and propose a framework that effectively exploits the relationship between semantic words for summarization. We achieve this goal by treating semantic words as nodes in a graph and learning their interactions by coupling Graph Convolutional Network (GCN) and Long Short Term Memory (LSTM), with the aid of visual cues. Two schemes of GCN-LSTM Interaction (GLI) modules are proposed for seamless integration of GCN and LSTM. The effectiveness of our approach is demonstrated via an extensive comparison with the state-of-the-arts methods on the two benchmarks ActivityNet Captions dataset and YouCook II dataset.
\end{abstract}


% \fi
\IEEEpeerreviewmaketitle

\section{Introduction}

\IEEEPARstart{D}{ense} video captioning, which aims at detecting all events and giving language descriptions in an untrimmed long video, is a very challenging problem in computer vision and has attracted a lot of research attentions recently.
This task consists of two sub-tasks: 1) temporal proposal generation to localize the events and 2) video captioning to describe the events. 

Existing dense video captioning methods have attempted to improve the solutions of both sub-tasks~\cite{krishna2017dense, li2018jointly,zhou2018end,wang2018bidirectional}. As for the temporal proposal generation sub-task, krishna et al.~\cite{krishna2017dense} employed the Deep Action Proposals (DAPs)~\cite{escorcia2016daps}, while Wang et al.~\cite{wang2018bidirectional} extended Single Stream Temporal Action Proposals (SST)~\cite{buch2017sst} to bidirectional Single Stream Temporal Action Proposals (bi-SST) to utilize both the past and future contexts. Li et al.~\cite{li2018jointly} and Zhou et al.~\cite{zhou2018end} both proposed anchor-based temporal proposal generation methods.  For the video captioning sub-task, the works in ~\cite{krishna2017dense, wang2018bidirectional, li2018jointly} used Long Short Term Memeory (LSTM) with attention mechanism as the decoder to generate sentences for event description, with the help of additional temporal information outside of the proposal. A recent work in~\cite{zhou2018end} proposed to apply a masking transformer network as the captioning decoder.

While these methods achieved promising results for dense video captioning, they all work at the whole proposal level to generate the descriptive sentence without sufficiently exploring the scene evolution within an event proposal. Therefore, they do not work well when the scenes and objects change largely over a relatively long proposal. 

In this paper, we focus on the second sub-task (i.e., video captioning). 
Observing that the generated proposal is still long and therefore the scenes and objects within it may rapidly change, we propose a graph-based partition-and-summarization (GPaS) framework to improve video captioning. In this framework, we generate semantic words at finer levels and exploit word relationships at different semantic levels to integrate the semantic words for event description. Specifically, in the ``partition" module, a temporal proposal is divided into multiple segments, from which both visual features and semantic sentences are generated to describe the temporally local information. In the ``summarization" module, the final sentence to describe the whole event proposal is generated by using both visual features and segment words. In this module, word relationships are explored by using the graph. 

In our GPaS framework, we use semantic words as the nodes in the graph, on top of which the Graph Convolutional Network(GCN)\cite{velivckovic2017graph,yang2018graph} highly interacts with LSTM to boost the sentence summarization performance. Specifically, LSTM provides GCN the features (hidden representations of words) that characterize each graph node, while GCN updates nodes to help LSTM to refine the hidden states based on the node (word) relationship. To seamlessly integrate GCN and LSTM, two versions of GCN-LSTM Interaction modules are proposed by placing GCN node either inside or outside a LSTM cell, respectively. GCN and LSTM can be jointly optimized in an end-to-end fasion to benefit from each other.

Specifically, in our graph-based summarization approach, we consider the node relationship based on two types of graphs. In the type I graph (referred as the basic graph), the input words are connected to the output words via directional edges in the summarization module. In the type II graph (referred as the expanded graph), the segment-level sentences are introduced as nodes in an additional layer between the input words and the output words in the summarization module, to explore their inner relationships for summarization. The input words, segment-level sentences, and output words in the summarization module are all treated as nodes at different semantic levels in the same graph. In our GPaS framework, we use GCN to learn the relationship of nodes across multiple semantic levels, while we use LSTM to learn the relationship of nodes within the same semantic level. 

The major \textbf{contributions} of our work are as follows. First, we propose a new graph-based partition and summarization (GPaS) approach to explore the fine details at the sub-proposal level for dense video captioning and treat both the input words and the output words in the summarization module as nodes of a graph (i.e., the basic graph). Secondly, we propose two types of GCN-LSTM Interaction (GLI) modules to couple GCN and LSTM seamlessly for sentence summarization. In addition, we further extend the basic graph into the expanded graph by introducing the segment-level sentences from the partition module as the additional nodes.  Third, our approach is evaluated on the two benchmarks: ActivityNet Captions dataset and YouCook II dataset, and compared widely with the state-of-the-arts dense video captioning methods, which demonstrates promising performance. 
\color{black}
\section{Related Work}

Dense video captioning is implemented based on two sub-tasks, temporal action proposal methods \cite{duchenne2009automatic,caba2016fast,escorcia2016daps,buch2017sst,wang2018bidirectional, su2019improving} and image/video captioning methods \cite{shen2017weakly,pan2016jointly,chen2018less,wang2018video,wang2018reconstruction,wu2018interpretable,chen2018regularizing, li2019know,yang2018multitask,zhang2019show}. In addition, our work is related to graph convolutional network \cite{kipf2016semi,yang2018graph,velivckovic2017graph,yao2018exploring,narasimhan2018out}. They are reviewed as follows.

\subsection{Temporal action proposals}

Temporal action/event proposals are temporal segments that possibly contain actions/events over a long untrimmed video.
The Deep Action Proposals (DAPs) method \cite{escorcia2016daps} is the first work that introduced the LSTM model to encode visual contents and generate action proposals within a sliding window. Based on DAPs~\cite{escorcia2016daps}, the Single-stream Temporal action proposals (SST) method was proposed in \cite{buch2017sst} to accelerate detection speed, which applied  a single processing stream instead of multiple sliding windows for proposal generation.  This method was further extended to the directional SST method in \cite{wang2018bidirectional}, where two directional processing streams were utilized to localize the action proposals by using past and future visual information.

\subsection{Image and video captioning}

The recent image/video captioning works are sequence learning approaches based on deep 
neural networks such as Convolutional Neural Networks (CNNs) and LSTM. Usually, CNNs are used as the encoder to encode images or video frames, while the LSTM networks are used as the decoder to generate descriptive sentences~\cite{li2018jointly}. Different frame encoding strategies have been proposed, such as mean-pooling \cite{venugopalan2014translating,gan2017semantic}, attention mechanism\cite{yao2015describing,pan2017video,gan2017semantic,zhang2018high,gao2017video,yan2019stat,tan2019comic} or recurrent networks \cite{sutskever2014sequence,donahue2015long,pan2016hierarchical,wu2019recall,xiao2019deep}. However, these video captioning methods are  not specifically designed for handling long videos with multiple events. In contrast, our work focuses on dense video captioning that generates a set of descriptive sentences for all events in each video \cite{krishna2017dense}. 

Some methods for addressing the visual and textual mapping problems are also related to our work. For example, in the works \cite{tang2016generalized} and \cite{shu2015weakly}, the author proposed to use the deep transform network to transfer the information between the visual domain and the textual domain and solve the textual and visual mapping problems with insufficient training examples. Their ideas can also be used in the image and video captioning problems when the training data is limited.

It is worth noting that the work in \cite{li2015summarization,liu2016boosting, zhang2019show} also used the ``partition and summarization"(PaS) strategy as in our approach for video captioning. The works in \cite{li2015summarization,liu2016boosting} used the pre-trained image captioning model at the frame-level to generate multiple sentences for a set of video key frames and then applied a simple summarization strategy for video captioning. For example, the work in \cite{li2015summarization} used LexRank \cite{erkan2004lexrank} to select the most representative segment-level sentence to describe the whole video. The work in \cite{liu2016boosting} employed a basic sequence-to-sequence architecture to integrate segment-level sentences into one sentence for video description. The work in \cite{zhang2019show} used the  hierarchical attention mechanism for the  summarization module. In contrast to these works\cite{li2015summarization,liu2016boosting,zhang2019show}, we propose a new sentence summarization framework by utilizing the GCN to model word relationship.

\subsection{Graph Convolutional Network and Recurrent Neural Network}
The work in \cite{kipf2016semi} proposed the Graph Convolutional Network (GCN) method for semi-supervised image classification by using the predefined adjacency matrix to refine the feature of each node based on other connected nodes. The work in \cite{velivckovic2017graph} further extended GCN into Graph Attentional Network, in which the adjacency matrix is learned by using attention mechanism. GCN has been applied to many computer vision tasks, such as scene graph generation \cite{yang2018graph}, visual question and answering\cite{narasimhan2018out}, and image captioning \cite{yao2018exploring}.

LSTM (as well as other recurrent neural networks) and GCN have been separately used in many research works. In \cite{veeriah2015differential}, differential recurrent neural network was proposed for human action recognition. In \cite{peng2019few}, a knowledge transfer network architecture was proposed, which incorporated visual feature learning, knowledge inferring and classifier learning for few-shot image recognition. These works are related to our proposed framework by using either the GCN-based networks or the modified RNN-based networks. However, the way they use GCN and RNN is significantly different from our framework. Each of these works improves either GCN or RNN models separately and none of them considers the integration of GCN and RNN networks. Particularly, in \cite{veeriah2015differential}, high-order derivatives of the RNN cell states were proposed to learn the spatial-temporal dynamics  of  various  actions. In \cite{peng2019few}, GCN is introduced to learn the semantic-visual mapping.  In contrast to these works, our proposed framework has two major differences. First, we tightly couple the GCN nodes and LSTM cells to refine the hidden representations/cell states. Second, we exploit different levels of semantic meanings of the sentences in the summarization module by using the proposed expanded graph via the GCN whose nodes are defined by the LSTM cells. 

In \cite{shu2019hierarchical}, a hierarchical LSTM architecture that contains concurrent LSTM units was proposed for human interaction recognition. This architecture can deal with the change of human interactions over time. In \cite{tang2019coherence}, the authors proposed a coherence constrained graph LSTM architecture that imposes spatial-temporal context coherence constraint and a global context coherence constraint for group activity recognition. In \cite{shu2019spatiotemporal}, a spatial-temporal co-attention recurrent neural network was proposed for human motion prediction. This network can deal with spatial coherence among joints and temporal evolution among skeletons. The works in \cite{shu2019hierarchical, shu2019spatiotemporal} focus on enhanced RNN-based networks without using graph convolutional network.  Our proposed GPaS framework is also intrinsically different from\cite{tang2019coherence} in the following aspects. First, we use one word as one node with the semantic feature as its corresponding feature for the dense video captioning task. The work in \cite{tang2019coherence} uses one person as one node with the visual feature as its corresponding information for the activity recognition task. In addition, we explore not only the sequential relationship among words, but also the hierarchical relationship across different semantic levels, where the hierarchical relationship is modeled by GCN. In contrast, the graph used in \cite{tang2019coherence} is not a hierarchical graph.

Using both GCN and LSTM, the work in \cite{yao2018exploring} is closely related to our work.  It employed faster-rcnn \cite{ren2015faster} for object detection and regarded the detected objects as nodes. The refined feature of each node after using GCN is directly fed into the LSTM-based decoder by using mean pooling/attention mechanism. Our approach is different from \cite{yao2018exploring} in at least two major aspects. First, from the conceptual perspective, we treat semantic words rather than the detected objects as in \cite{yao2018exploring} as nodes and exploit the semantic relation rather than the visual relation. Second, from the architecture perspective, our approach tightly couples the GCN nodes with the LSTM cells, while in \cite{yao2018exploring} the GCN-based encoder and the LSTM-based decoder were treated as two separate modules, which are simply concatenated together.

\section{Methodology}
\begin{figure}[t]
    \centering
    \includegraphics[width=\linewidth]{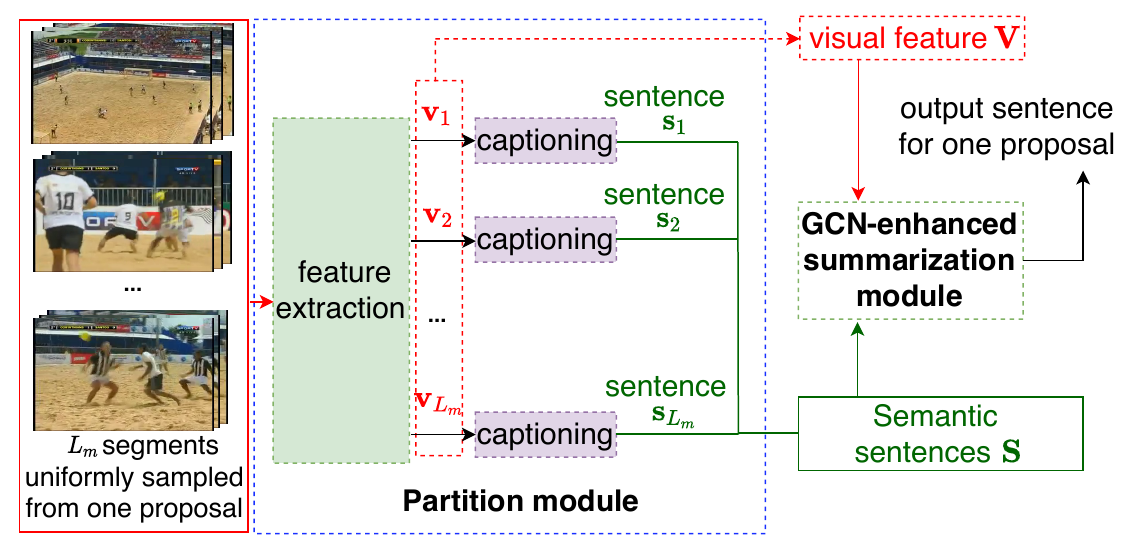}
    \caption{Overview of our GPaS framework. Our framework consists of a partition module, which generates multiple sentences from the sampled frames,  and a summarization module, which summarizes the sentences through GCN-LSTM Interaction.}
    \label{overview}
\end{figure}

Fig. \ref{overview} shows an overview of our GPaS framework for dense video captioning. Given a long untrimmed video, we utilize the Bi-AFCG method \cite{wang2018bidirectional} to generate temporal proposals for possible actions in the video. To generate sentence description for each proposal, our framework consists of a partition module and a summarization module.  We follow the existing work \cite{zhang2019show} to split each video proposal into multiple segments, extract visual features and generate textual description for each segment (see Section \ref{Sec:Division} for more details). The summarization module, with the details in Section~\ref{Sec:summarisation}, integrates the textual descriptions and visual features from all video segments to generate the overall sentence for the event description. This work focuses on the summarization module and we will elaborate our proposed GCN-enhanced summarisation framework in Section~\ref{Sec: GCN-enhanced-module}.

\subsection{Partition module}\label{Sec:Division}

  To fully explore the rich content inside one video proposal, we follow the work in \cite{zhang2019show} to partition one proposal into several video segments and extract the visual and textual features from each segment for better description. Specifically, a predetermined number (i.e., $L_m=20$) of video segments are uniformly sampled from the temporal range determined by the temporal proposal. After sampling, each segment has 16 frames. For the $j$-th segment , the 3D visual feature  ${\mathbf v}_j$ is extracted by using the 3D feature extractor. Based on these visual features, sentence description is then generated for each segment by applying the captioning method in~\cite{vinyals2015show}. In this way, the partition module outputs two kinds of information, i.e., the visual feature ${\mathbf V} = \{{\mathbf v}_1, {\mathbf v}_2, \cdots, {\mathbf v}_{L_m} \}$, and the segment-level sentences ${\mathbf S} = \{{\mathbf s}_1, {\mathbf s}_2, ..., {\mathbf s}_{L_m} \}$. Each segment-level sentence ${\mathbf s}_j$ contains $L_k$ words, ${\mathbf s}_j = \{{\mathbf w}_{(j-1) \times L_k+1}^{in}, \cdots, {\mathbf w}_{j \times L_k}^{in} \}, j = 1,2,...,L_m$, where ${\mathbf w}^{in}_t$ is the $t$-th output word for partition module (i.e., the $t$-th input word for summarization module) in ${\mathbf S}$ and superscript $in$ in ${\mathbf w}^{in}_t$ denotes the input word for the summarization module.

After performing the captioning method in \cite{vinyals2015show}, we would like to mention that different segments may be described by different numbers of words. In order to facilitate the training process, our method assumes each segment has the same number of words $L_k$. To this end,  we adopt a post-processing method. Specifically, if the sentence is longer than $L_k$ words, only the first $L_k$ words are kept; if the sentence is shorter than $L_k$ words,  the signal $<blank>$ is added after the period of sentence until the sentence has $L_k$ words.

\subsection{Basic summarization module}\label{Sec:summarisation}

\begin{figure}[t]
    \centering
    \includegraphics[scale = 0.62]{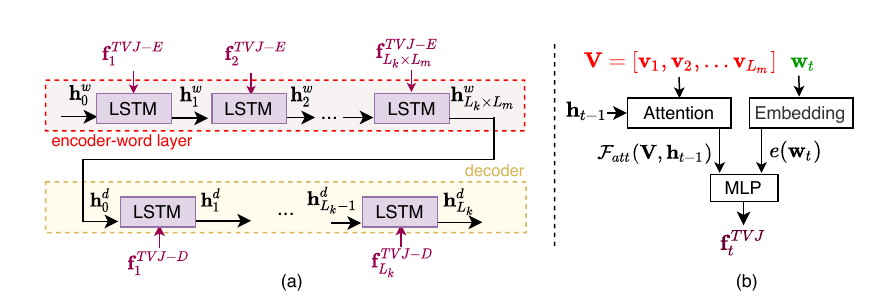}
    \caption{(a)Architecture of the basic summarization module. The red block represents the encoder-word layer while the yellow block represents the decoder. (b) General concept of the textual feature and visual feature joint module (TVJ). For the encoder,  we replace $\mathbf{h}_{t-1}$ and $\mathbf{w}_{t}$ by using $\mathbf{h}_{t-1}^w$ and $\mathbf{w}^{in}_{t}$  respectively,  which represent the hidden state in the encoder-word layer and the output words from the partition module (i.e., the input words for summarization module), respectively. For the decoder,  we replace $\mathbf{h}_{t-1}$ and $\mathbf{w}_{t}$ by using $\mathbf{h}_{t-1}^d$ and $\mathbf{w}^o_{t-1}$ respectively, which represent the hidden state in the decoder and the generated output words in the last time step, respectively. \wl{$\mathbf{w}^{in}_{t}$ and $\mathbf{w}^o_{t-1}$ are the notations of the “input word” and “output word” in the summarization module, respectively.}} 
    \label{basic}
\end{figure}

Our basic summarization module follows the existing work \cite{zhang2019show} by adopting the “encoder-decoder” structure shown in Fig. \ref{basic} (a). The red block refers to the encoder-word layer  while the yellow block refers to the decoder.

The encoder-word layer generates the hidden representations for each input word obtained in the partition module, and the decoder used the last hidden representations from the encoder-word layer to generate one-sentence description for each event proposal. 

\subsubsection{Basic Unit}
As in \cite{zhang2019show}, both the encoder-word layer and the decoder are built upon the basic units comprising of a LSTM cell with $\mathbf{f}^{TVJ}_t$ as the input, where $\mathbf{f}^{TVJ}_t$ is the output of the textual feature and visual feature joint module (\textbf{TVJ}). The basic units are stacked to form the encoder-word layer or the decoder. Different from the common sequence models used in the summarization module such as in \cite{liu2016boosting}, where only semantic words are used to generate the hidden representation, the input of our LSTM cell is the output of the \textbf{TVJ} module according to Eqn.~(\ref{Eqn:VTF-fusion}):
\begin{equation}\label{Eqn:VTF-fusion}
\mathbf{f}^{TVJ}_t = \mbox{MLP}(f_{att}({\mathbf V},{\mathbf h}_{t-1}), e({\mathbf w}_{t})),
\end{equation}
where $t$ indicates the $t$-th time step in LSTM or equally the index of the $t$-th input word ${\mathbf w}_{t}$; the symbol ${\mathbf V}$ denotes the visual features extracted from all segments within one proposal in the partition module, and ${\mathbf h}_{t-1}$ indicates the LSTM hidden state at $t-1$ time step for either the encoder-word layer or the decoder. $e(\cdot)$ is a word embedding function to transform a one-hot semantic word into a feature vector. $\mbox{MLP}(\cdot)$ is a multi-layer perceptron for fusing the visual and word features. The attention function $\mathcal{F}_{att}({\mathbf V},{\mathbf h}_{t-1})$ combines the visual features based on their importance:
\begin{align}\label{Eqn:f-att}
f_{att}({\mathbf V}, {\mathbf h}_{t-1}) = \sum_{i=1}^{L_m}a_{t,i}*{\mathbf v}_i,   
\end{align}

where ${\mathbf v}_i$ indicates the $i$-th feature of $\mathbf{V}$ and $L_m$ the total number of visual features. $a_{t,i}$ in Eqn.(\ref{Eqn:f-att}) is the $i$-th element of the vector ${\mathbf a}_{t}$, where ${\mathbf a}_{t} = softmax (\mbox{MLP}({\mathbf V}, {\mathbf h}_{t-1}))$  and $\mbox{MLP}(\cdot)$ is the multi-layer perceptron that takes the concatenated vectors ${\mathbf V}$ and ${\mathbf h}_{t-1}$ as the input. With $\mbox{MLP}(\cdot)$ and $softmax$, the learnable weights ${\mathbf a}_{t}$ is obtained. 

Then in each LSTM cell, the hidden representation of the $t$-th word is generated by ${\mathbf h}_{t} = LSTMCell( \mathbf{f}^{TVJ}_t,  \textbf{h}_{{t}-1})$ in the normal way as in ~\cite{hochreiter1997long}, which is also illustrated in Fig.~\ref{fig:aGCN-LSTM}(a).

The formulations and symbols above, e.g. in Eqn.(\ref{Eqn:VTF-fusion}) and Eqn.(\ref{Eqn:f-att}), are general representations that will be replaced by the corresponding symbols of the encoder-word layers and decoders below. 

\subsubsection{Implementing Encoder-word layer and Decoder using the Basic Unit}
\textbf{Encoder-word layer}~~~~Our encoder-word layer comprises of $L_m \times L_k$ basic units for $L_k$ words per segment and $L_m$ segments, each basic unit corresponding to one segment word. 
The input words and visual features are fused in the TVJ module. The output of the TVJ module in the encoder-word layer $\mathbf{f}^{TVJ-E}_t$ in (\ref{Eqn:VTF-fusion}) is modified as $ \mbox{MLP}(f_{att}({\mathbf V},{\mathbf h}_{t-1}^w), e({\mathbf w}^{in}_{t}))$, where ${\mathbf h}_{t}^w$ is the hidden representation in the encoder, ${\mathbf h}_{t}^w=LSTMCell( \mathbf{f}^{TVJ-E}_t,  \textbf{h}_{{t}-1}^w)$. Here $\mathbf{w}^{in}_t$ is the $i$-th word generated by the partition module introduced in Section \ref{Sec:Division}, and $\mathbf{V}$ is the visual input.
~The output of the encoder-word layer is the last hidden representation ${\mathbf h}_{L_k \times  L_m}^w$, which is fed into the decoder of the basic summarization module.

\noindent\textbf{Decoder}~~~~ The decoder takes the hidden state in the previous time step and the visual features ${\mathbf V}$ as the input, and comprises of $L_k$ basic units with each corresponding to an output word constituting the final sentence description. The initial hidden state $\mathbf{h}^d_0$ of the decoder is initialized from the last hidden representation of the encoder-word layer ${\mathbf h}^w_{L_m \times L_k}$.
Similar to the encoder-word layer, the basic units in the decoder output
${\mathbf h}^{d}_{t} = LSTMCell( \mathbf{f}^{TVJ-D}_t,  {\mathbf h}^{d}_{{t}-1})$ at the time step $t$. Here $\mathbf{h}^d_{{t}}$ is the hidden state of the decoder at time step $t$.  The output word ${\mathbf w}_t^o$ in the final sentence description is obtained by learning a transformation matrix that maps a hidden state $\mathbf{h}^d_{{t}}$ to the probabilities of the words in the dictionary. In the decoder, $\mathbf{f}^{TVJ-D}_t$ is computed by replacing ${\mathbf h}_{t-1}$ with ${\mathbf h}^d_{t-1}$ and replacing ${\mathbf w}_t$ with the word ${\mathbf w}_{t-1}^o$ in Eqn.(\ref{Eqn:VTF-fusion}) in the testing stage.

\begin{figure*}[t]
    \centering
    \includegraphics[width=\linewidth]{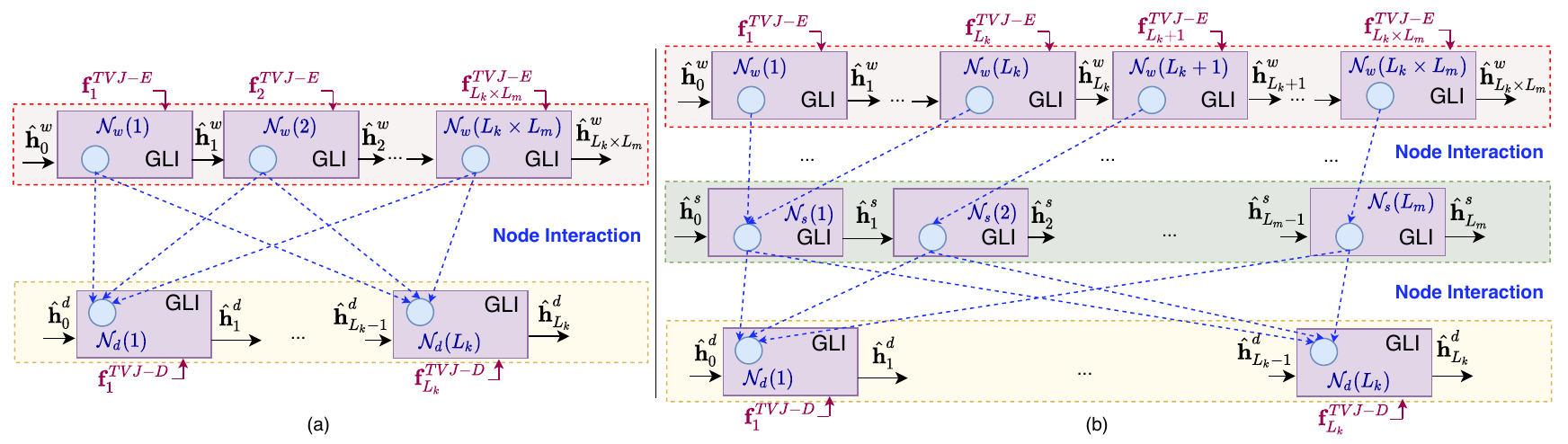}
    \caption{ Architecture of our GCN-enhanced summarization module with (a) with two semantic levels and (b) with three semantic levels. The red block indicates the encoder-word layer, the green block indicates the encoder-segment layer and the yellow block indicates the decoder. The blue circles represent the nodes in the GLI modules. They are connected by the dashed lines to constitute (a) the basic graph and (b) the expanded graph. } 
    \label{fig:one-graph}
\end{figure*}

\subsection{GCN-enhanced summarization module}\label{Sec: GCN-enhanced-module}
In our GPaS framework, the quality of the integrated input words in the summarization module directly affects the quality of the one-sentence description for an event. 
Therefore, based on our basic summarization module, in order to explore the relationship between semantic words, we treat both the input words in the encoder-word layer and the generated output words in the decoder as the nodes in a graph (called the basic graph). The semantic features of these words are regarded as the corresponding features of these nodes. To further investigate the importance of each segment, we introduce $L_m$ sentences from $L_m$ segments as the additional nodes in an expanded graph. 

\noindent \textbf{Attentional GCN}
Graph convolutional network (GCN) \cite{velivckovic2017graph} builds upon a directed graph connecting the nodes characterized by the node features. It simplifies the computation over the graph by updating the node features based on the adjacent nodes, which are weighted by their predetermined connection strength. Attentional GCN (aGCN)~\cite{yang2018graph} further makes the node connection strength learnable rather than predetermined. Let ${\mathcal{N}}(i)$ denote the $i$-th node in the graph with the corresponding feature vector ${\mathbf z}_i$ at the initial stage, and let ${{\mathcal P}_i}$ denote the index set of the nodes that point to ${\mathcal{N}}(i)$ with direct connection. With one round of node update in aGCN, ${\mathbf z}_i$ becomes ${\hat{\mathbf z}}_i$ based on the following Eqn.(\ref{Eqn:aGCN}):
\begin{equation}
{\hat{\mathbf z}}_i = tanh({\mathbf z}_i+ {\mathbf W}\sum_{j \in  \mathcal{P}_i} \alpha_{ij}{\mathbf z}_j).
\label{Eqn:aGCN}
\end{equation}

Here ${\mathbf z}_j$ denotes the initial feature vector of the node ${\mathcal N}(j)$ with $j \in  {\mathcal P}_i$. $\alpha_{ij}$ provides the learnable weight for the information passing from ${\mathcal{N}}(j)$ to ${\mathcal{N}}(i)$, and is defined as:
\begin{equation}
\begin{split}
 u_{ij} &= MLP([{\mathbf z}_i, {\mathbf z}_j])  \\ 
{\boldsymbol \alpha}_{i} &= softmax({\mathbf u}_{i}). 
\end{split}
\label{eq:gcn_att}
\end{equation}

Here ${\boldsymbol \alpha}_{i}$ is a vector with its $j$-th element as $\alpha_{ij}$, containing the learnable weights. ${\mathbf u}_{i}$ is a vector with its $j$-th element as ${u}_{ij}$. To distinguish different message passing stages, a superscript is often used with $\alpha_{ij}$ to indicate its specific representation. MLP is a two-layer preceptron. Only the nodes that point to ${\mathcal{N}}(i)$ pass information to ${\mathcal{N}}(i)$. 
Otherwise, $\alpha_{ij} = 0, \forall j \not\in \mathcal{P}_i$.\\
%Otherwise, $u_{ij} = 0, \forall j \not\in \mathcal{P}_i$, and accordingly $ \alpha_{ij} = 0, \forall j \not\in \mathcal{P}_i$.\\

\noindent \textbf{The proposed aGCN setting}~~In our summarisation module, aGCN is used to explore the relationship of words across semantic levels to refine their features, as shown in Fig.~\ref{fig:one-graph} (a) and Fig.~\ref{fig:one-graph} (b).  We propose a GLI module consisting of one GCN node and a LSTM cell, illustrated in Fig.~\ref{fig:aGCN-LSTM} (b) and Fig.~\ref{fig:aGCN-LSTM} (c). The GLI module has 1) the function of LSTM, which can deal with sequential information and 2) the function of aGCN, which can refine the features of the nodes from a graph. The nodes from the GLI modules can constitute a graph. Two graphs with different structures are proposed for this purpose. 

The first graph, referred as the \textbf{basic graph}, is modeled in Fig.~\ref{fig:one-graph} (a), which is based on the two-layer (the encoder-word layer and the decoder) GLI. There are two sets of graph nodes. The nodes in the first set represent the $L_m \times L_k$ input words in the encoder-word layer (denoted as the set ${\mathcal N}_{w}$). The nodes in the second set represent the $L_k$ output words in the decoder (denoted as the set ${\mathcal N}_{d}$). The nodes in  ${\mathcal N}_{w}$ are connected with the nodes in ${\mathcal N}_{d}$ along the direction from ${\mathcal N}_{w}$ to ${\mathcal N}_{d}$, and the node features are updated by following the rule of aGCN in Eqn.(\ref{Eqn:aGCN}).

The second graph, referred as the \textbf{expanded graph}, is modeled in Fig.~\ref{fig:one-graph} (b). It is proposed to expand our two-layer GLI into three-layers (the encoder-word layer, the encoder-segment layer and the decoder) GLI for summarization. That is, we insert the encoder-segment layer, corresponding to the $L_m$ segment-level sentences. Accordingly, in the expanded graph,  
a third set of nodes are introduced in the encoder-segment layer, i.e., the nodes representing the $L_m$ video segment-level sentences, are denoted as the set ${\mathcal N}_{s}$. 
As shown, the nodes in ${\mathcal N}_{w}$ corresponding to the input words in the summarization module are connected to the node in ${\mathcal N}_{s}$, which represents their corresponding segment-level sentence. Then the nodes in ${\mathcal N}_{s}$ are further connected to the nodes in ${\mathcal N}_{d}$. Again, the node features are updated based on Eqn.(\ref{Eqn:aGCN}).

From Fig.~\ref{fig:one-graph} (a) and Fig.~\ref{fig:one-graph} (b), the aGCN part from the GLI module is used to model the relationship of words across different semantic levels (input words in the encoder-word layer, segment-level sentences in the encoder-segment layer and output words in the decoder). For the words within each semantic level, their relationship is modeled by the LSTM part of GLI module. In the following, we propose two types of the GLI module for summarization.\\

\subsection{GCN outside LSTM} 
\begin{figure*}[t]
    \centering
    \includegraphics[scale = 0.81]{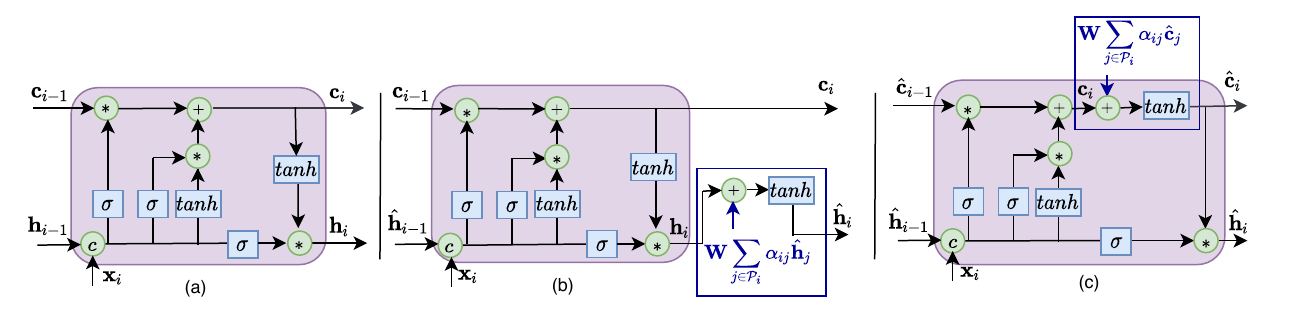}
    \caption{Illustration of (a) the basic LSTM cell and two types of GLI modules (b) aGCN-out-LSTM and (c) aGCN-in-LSTM. The blue block represents the update from an aGCN node by refining the hidden state in (b) and the cell state in (c).}
    \label{fig:aGCN-LSTM}
\end{figure*}
The first type of our GLI module is called aGCN-out-LSTM, where the node of aGCN is placed outside of the LSTM cell, as illustrated in Fig.~\ref{fig:aGCN-LSTM} (b). In this framework, the hidden state of each LSTM cell is used as the graph node feature in aGCN. Therefore, these hidden states could be refined via the mechanism of aGCN according to Eqn.(\ref{Eqn:aGCN}). We adopt a progressive refinement strategy, that is, the refinement of the nodes at each semantic level depends on the updated features of the nodes in its preceding level. 

\noindent\textbf{Basic graph}~~For each node ${\mathcal N}_w(i) \in {\mathcal N}_w$ corresponding to an input word, in the encoder-word layer, its initial feature is set as ${\mathbf h}^w_i$. Since there are no nodes pointing to ${\mathcal N}_w(i)$, i.e., ${\alpha}_{ij}=0$, the refined node feature $\hat{\mathbf z}^w_i$, or equally the hidden state $\hat{\mathbf h}^w_i$ of the corresponding LSTM cell, does not change after updating each aGCN, which is 
\begin{equation}\label{Eqn:basic-graph-NeUpdate}
\begin{split}
 \hat{\mathbf h}^w_i &= {\mathbf h}^w_i. 
\end{split}
\end{equation}

For each node ${\mathcal N}_d(i) \in {\mathcal N}_d$ corresponding to an output word in the decoder, the initial node feature is set as ${\mathbf h}^d_i$. Pointed by $L_k \times L_m$ nodes in ${\mathcal N}_w$, the feature of ${\mathcal N}_d(i)$ is updated by:
\begin{equation}
 \hat{\mathbf h}^d_i = tanh({\mathbf h}^d_{i} + {\mathbf W}^{d} \sum_{j=1}^{L_k \times L_m} \alpha_{ij}^{d}  \hat{\mathbf h}^w_j).  
\end{equation}
where $\mathbf{h}^{d}_i$ is the hidden state in the decoder at time step $i$ and $\hat{\mathbf h}^w_j$ is the refined hidden state in the encoder-word layer at time step $j$. $\mathbf{W}^{d}$ is the learnable matrix while $\alpha_{ij}^{d}$ is the learnable weight defined in Eqn.(\ref{eq:gcn_att}).

\noindent\textbf{Expanded graph}~~The feature update process for each node ${\mathcal N}_w(i) \in {\mathcal N}_w$ is the same as Eqn.(\ref{Eqn:basic-graph-NeUpdate}) in the basic graph.

For each node ${\mathcal N}_s (i) \in {\mathcal N}_s$ corresponding to a segment-level sentence which is from the encoder-segment layer, we use the hidden state of the corresponding LSTM cell ${\mathbf h}_i^s$ as the initial node feature, the refined feature of ${\mathcal N}_s(i)$,  or equally the refined hidden state $\hat{\mathbf h}_i^s$, is updated by:
\begin{equation}
\begin{split}
 \hat{\mathbf h}^s_i = tanh({\mathbf h}^s_{i} +  \tilde{\mathbf{W}}^s \sum_{j=L_k \times (i-1)+1}^{L_k \times i} \tilde{\alpha}_{ij}^s \hat{\mathbf h}^w_j). 
\end{split}
\end{equation}
Here $\mathbf{h}^s_i$ is the hidden state in the encoder-segment layer at time step $i$ and $\hat{\mathbf h}^w_j$ is the refined hidden state in the encoder-word layer at time step $j$. $\tilde{\mathbf{W}}^{s}$ is the learnable matrix while   $\tilde{\alpha}_{ij}^{s}$ is the learnable weight defined in Eqn.(\ref{eq:gcn_att}).
Finally the update of the node feature  for ${\mathcal N}_d(i) \in {\mathcal N}_d$ from the decoder, or equally the  refined hidden state $\hat{\mathbf h}_i^d$, becomes:
\begin{equation}
\begin{split}
 \hat{\mathbf h}^d_i = tanh({\mathbf h}^d_{i} + {\tilde{\mathbf W}}^{d} \sum_{j=1}^{L_m} \tilde{\alpha}_{ij}^{d} \hat{\mathbf h}^s_j).  
\end{split}
\end{equation}
Here $\mathbf{h}^d_i$ is the hidden state in the decoder at time step $i$ and $\hat{\mathbf h}^s_j$ is the refined hidden state in the encoder-segment layer at time step $j$. $\tilde{\mathbf{W}}^{d}$ is the learnable matrix while $\tilde{\alpha}_{ij}^{d}$ is the learnable weight defined in Eqn.(\ref{eq:gcn_att}).

As shown, with aGCN, the hidden states of the LSTM cells could be refined by those from the other semantic levels. Through the end-to-end training process, aGCN and LSTM can collaborate with each other in the GLI module to improve the summarization result.

\subsection{GCN inside LSTM}

In our second type of GLI module, called aGCN-in-LSTM, the node of aGCN is placed inside the LSTM cell, as illustrated in Fig.~\ref{fig:aGCN-LSTM} (c). This time, the node feature of aGCN is governed by the LSTM cell state ${\mathbf c}_t$. 

Recall that in a standard LSTM cell in Fig.~\ref{fig:aGCN-LSTM} (a), the memory cell ${\mathbf c}_t$ and the hidden state ${\mathbf h}_t$ are obtained as ${\mathbf c}_t = {\mathbf f}_t \odot {\mathbf c}_{t-1} + {\mathbf i}_t \odot {\mathbf g}_t$, and ${\mathbf h}_t = {\mathbf o}_t \odot \tanh({\mathbf c}_t)$, where ${\mathbf f}_t$, ${\mathbf i}_t$ and ${\mathbf o}_t$
correspond to the forget gate, the input gate and the output gate, respectively. The operator $\odot$ denotes the element-wise multiplication. Therefore, the refinement of ${\mathbf c}_t$ leads to the refinement of ${\mathbf h}_t$.

Specifically, the cell state of a LSTM cell corresponding to the node ${\mathcal N}(i)$ in aGCN is refined as:
\begin{equation}
\begin{split}
\hat{\mathbf c}_i &= tanh({\mathbf c}_i + {\mathbf W}\sum_{j \in {\mathcal P}_i} \alpha_{ij} \hat{\mathbf c}_j),  \\
\hat{\mathbf h}_i &= {\mathbf o}_i \odot \hat{\mathbf c}_i.
\end{split}
\label{eq:aGCN_cell_inside}
\end{equation}
Recall that ${\mathcal P}_i$ indicates the index set of nodes pointing to ${\mathcal N}(i)$, as mentioned in Eqn.(\ref{Eqn:aGCN}). 

Based on Eqn.(\ref{eq:aGCN_cell_inside}), it is not difficult to derive the refinement of nodes in ${\mathcal N}_w$, ${\mathcal N}_s$ and ${\mathcal N}_d$, in a similar way as for aGCN-out-LSTM. To avoid repetition, we skip the details and directly give the update of ${\mathbf c}_i$ based on the expanded graph as follows.
\begin{equation}
\begin{split}
& \hat{\mathbf c}^w_i = {\mathbf c}^w_i, \\
&\hat{\mathbf c}^s_i = tanh({\mathbf c}^s_i + \bar{\mathbf W}^{s} \sum_{j=L_k \times (i-1)+1}^{L_k \times i} \bar{\alpha}_{ij}^{s} \hat{\mathbf c}^w_j), \\ 
& \hat{\mathbf c}^d_i = tanh({\mathbf c}^d_{i} + \bar{\mathbf W}^{d} \sum_{j=1}^{L_m} \bar{\alpha}_{ij}^{d} \hat{\mathbf c}^s_j), 
\end{split}
\label{Eqn:aGCN_cell_inside_allnodes}
\end{equation}
where $\mathbf{c}^w_i$ and $\hat{\mathbf{c}}^w_i$ are the initial and refined cell states in the encoder-word layer at time step $i$, ${\mathbf c}^s_i$ and $\hat{\mathbf{c}}^s_i$ are the initial and refined cell states in the encoder-segment layer at time step $i$, and ${\mathbf c}^d_i$ and $\hat{\mathbf{c}}^d_i$ are the initial and refined cell states in the decoder at time step $i$.  $\bar{\mathbf{W}}^{s}$ and $\bar{\mathbf{W}}^{d}$ are the learnable matrices while $\bar{\alpha}_{ij}^{s}$ and $\bar{\alpha}_{ij}^{d}$ are the learnable weights defined in Eqn.(\ref{eq:gcn_att}).

Consequently, the corresponding hidden states $\hat{\mathbf{h}}_i^w$, $\hat{\mathbf{h}}_i^s$, and $\hat{\mathbf{h}}_i^d$ are updated by using $\hat{\mathbf{h}}_i^* = \mathbf{o}_i^*\odot \hat{\mathbf{c}}_i^*$, where $*$ can be $w, s$, or $d$.

\subsection{Other Details}
As mentioned, once the hidden representation $\hat{\mathbf h}_t^d$ of the decoder is learned, it needs to be mapped to a word in the final sentence. This is performed by optimizing a cross-entropy loss to estimate the probabilities of the words that $\hat{\mathbf h}_t^d$ corresponds to in a dictionary. In order to improve the prediction of words, we introduce a discriminative loss in addition to the cross-entropy loss, as follows:
\begin{equation}\label{Eqn:combined-cross-entropy-loss}
    \mathcal{L} = \mathcal{L}_{cross-entropy} +\lambda_{d} \mathcal{L}_{d},
\end{equation}
where $\mathcal{L}_{corss-entropy}$ is the cross entropy loss, and $\mathcal{L}_{d}$ is the discriminative loss \cite{yang2016review}. The hyper-parameter $\lambda_d$ is used to balance the two losses, and is empirically set to 0.1 in our experiment. Moreover,  we also apply  reinforcement learning \cite{rennie2017self} to boost our performance by using Meteor as the reward. 

\subsection {Difference from Attention}
Our approach has at least two main differences from the commonly used attention mechanism for word summarization. First, unlike the regular soft attention mechanism using the weighted sum attention on the input features  to  LSTM, our  attention  mechanism  in  GCN-LSTM is  used  to  refine  the  cell/  hidden  states  inside/outside  the LSTM  cells.   As  a  result,  we  introduce  more non-linearity,  which  improves  the  performance. Second, based
on our GCN model, multiple iterative refinements to the graph
nodes can be readily achieved.

\section{Experiments}\label{Sec:Experiments}

\subsection{Datasets}

As a dense video captioning benchmark, the ActivityNet Captions dataset~\cite{krishna2017dense} is used in our experiment. It contains around $20,000$ untrimmed long videos with temporally annotated sentences. Each sentence describes the event occurred in a corresponding video segment and contains $13.5$ words on average. Most videos are annotated with more than three sentences, and there are around $100,000$ sentences in total in the dataset. 

To evaluate our proposed model in longer videos, we also use an extra video captioning dataset called YouCook II. YouCook II has around 2,000 videos and the average length of video is more than 5 minutes, which is longer than the ActivityNet Captions dataset. For fair comparison,  we use the output of the ‘flatten-673’ layer in Resnet-200 as the appearance features and follow the work in \cite{zhou2018end} to extract the optical flow features. Only the validation set is used for performance evaluation as in \cite{zhou2018end}.

For data preprocessing, we discard all non-alphabetic characters and convert all the characters to lowercase. In addition, the tag $<UNK>$ will replace all the sparse words occuring less than three times in the training set, which eventually leads to a dictionary of 6,994 words on the ActivityNet Captions dataset and 1,269 words on the  YouCook II dataset.

\subsection{Experimental Setup}

In our experiment, for each proposal, the number of video segments (i.e., the number of segment-level sentences) is set as $L_m=20$. The maximum length of words for each sentence (i.e., the segment-level sentence or the final sentence for one event) is set as $L_k=25$. If the sentence is longer than 25 words, only the first 25 words are kept. If the sentence is shorter than 25 word, zero padding is applied. The total number of the input words in our summarization module is fixed as $500$.

Our model is implemented based on PyTorch. we use Adam~\cite{kingma2014adam} as the optimizer and the minibatch size is set to $32$.
The learning rate for the cross entropy loss is initially set to 0.0003 and then decreased 1.25 times every 3 epochs. The size of hidden states is $512$  and the dropout rate is $0.8$.

\subsection{Baselines and Evaluation Metrics}
\textbf{Baselines.}
We use the following state-of-the-arts methods as the baselines for performance comparison with our proposed approach.
(1) \textbf{LSTM} \cite{venugopalan2014translating} uses the LSTM model as the decoder. (2) \textbf{S2VT} \cite{venugopalan2015sequence} uses optical flow as the additional visual input. (3) \textbf{TA} \cite{yao2015describing} uses the attention mechanism to focus on the key part of the visual input. (4) \textbf{H-RNN} \cite{yu2016video} describes a video by using multiple sentences with a hierarchical LSTM. (5) \textbf{DCE} \cite{krishna2017dense} explores the information from the neighbouring event by using the attention mechanism. (6) \textbf{DVC}\cite{li2018jointly} proposes a temporal attention method to further boost the performance by employing attributes and reinforcement learning. (7) \textbf{Bi-AFCG}\cite{wang2018bidirectional} uses attention-based LSTM associated with context gating and joint ranking. (8) \textbf{MT} \cite{zhou2018end} uses mask transformer for proposal generation and captioning.

\textbf{Evaluation Metrics.} 
Four commonly used evaluation metrics are employed to measure the video captioning performance: 
~BLEU \cite{papineni2002bleu}, 
~Rouge-L\cite{lin2004rouge},
~Meteor \cite{denkowski2014meteor} and CIDEr-D \cite{vedantam2015cider}, in which Meteor is regarded as the most important indicator. When the proposals are automatically generated, we evaluated different methods based on different intersection over union (IoU) thresholds (0.3, 0.5, 0.7 and 0.9) and we report the mean results.
In the following tables, ~BLEU 1-4 \cite{papineni2002bleu}, 
~Rouge-L\cite{lin2004rouge},
~Meteor \cite{denkowski2014meteor} and CIDEr-D are simplified as B@1-B@4, R, M and C respectively.

\subsection{Performance Comparison}

\begin{table}[ht]
\centering
{
\caption{Performance comparison on the ActivityNet captions validation dataset when using the ground-truth proposals. R, B@4, M and  C refer to Rouge-L, BLEU4, Meteor and CIDEr-D respectively. }\label{Table:GT}
\begin{tabular}{|c|c|c|c|c|}
\hline
Model & R  & B@4  & M & C \\
\hline
LSTM \cite{venugopalan2014translating}   &- & 1.53 &8.66 & 24.07\\
\hline
S2VT \cite{venugopalan2015sequence}&-  &1.57   & 8.74 & 24.05 \\
\hline
TA \cite{yao2015describing} &-  & 1.56  & 8.75  &  24.14\\
\hline
H-RNN \cite{yu2016video}&-  & 1.59  & 8.81 & 24.17\\
\hline
DCE \cite{krishna2017dense}&-  & 1.60  &8.88 & 25.12 \\
\hline
DVC \cite{li2018jointly}&-  & \textbf{1.62}  & 10.33 & 25.24\\
\hline
GPaS (ours) & \textbf{21.30} & 1.53 & \textbf{11.04} & \textbf{28.20}\\
\hline
\end{tabular}
}
\end{table}

\begin{table}[!ht]

\centering
\caption{Performance comparison on the ActivityNet Captions validation dataset when using automatically generated  proposals. From the second to the fourth row, the results are based on the old evaluation criterion while the rest results are based on the new evaluation criterion.} 
\label{table:learned}
{
\begin{tabular}{|c|c|c|c|c|c|c|c|}
\hline
 Model & B@1 & B@2 & B@3 & B@4   & M & C \\
\hline
MT \cite{zhou2018end}& -&-& 4.76 & 2.23  & 9.56 & -- \\
\hline
Bi-AFCG \cite{wang2018bidirectional}& 18.99 & 8.84  &4.41 & 2.30  & 9.60 & 12.68 \\ \hline
GPaS (ours) & \textbf{19.78} & \textbf{9.96} & \textbf{5.06} & \textbf{2.34}  & \textbf{10.75} & \textbf{14.84} \\
\hline
\hline
 LSTM \cite{venugopalan2014translating}  & 11.19 & 4.73 & 1.75 & 0.60 & 5.53 & 12.06 \\
\hline
 S2VT \cite{venugopalan2015sequence} & 11.10 & 4.68 & 1.83 & 0.65  & 5.56 & 12.16 \\
\hline
 TA \cite{yao2015describing}   & 11.06 & 4.66 & 1.78 & 0.65  & 5.62 & 12.19 \\
\hline
 H-RNN \cite{yu2016video}  & 11.21 & 4.79 & 1.90 & 0.70 & 5.68 & 12.35 \\
\hline
 DCE \cite{krishna2017dense} & 10.81 & 4.57 & 1.90 & 0.71   & 5.69 & 12.43 \\
\hline
 DVC\cite{li2018jointly}   & 12.22 & 5.72 & 2.27 & 0.73  & 6.93 & 12.61\\
\hline
GPaS (ours)  & \textbf{13.73} & \textbf{6.02} & \textbf{2.54} & \textbf{0.93}  &  \textbf{7.44} & \textbf{13.00} \\
\hline
\end{tabular}
}
\end{table}

The performance of dense video captioning depends on both event proposal localization and caption generation. Since our work focuses on the latter, we first use the ground-truth proposals for performance evaluation in order to focus on  performance comparison on the validation set without considering the influence of proposals. For fair comparison, we follow most methods in this field to use the C3D features, and only compare our work with those baselines by using the same type of features in Table~\ref{Table:GT}. As shown in Table~\ref{Table:GT}, when the ground-truth proposals and C3D features are used, our proposed model \textbf{GPaS} achieves the highest Meteor score of $11.04\%$. It outperforms the second best performer~\textbf{DVC}~\cite{li2018jointly} by 0.71 \%, without using the additional action category labels as DVC does.  Our \textbf{GPaS} method also performs the best according to CIDEr-D. 

In Table~\ref{table:learned}, we also report the results on the ActivityNet Captions validation set by using the  automatically generated proposals. As the work in \cite{zhou2018end} uses the more advanced features, the P3D features is used in our proposed method in this experiment. The results in 
Table~\ref{table:learned} are partitioned into two parts according to two versions of evaluation criteria used in the literature. The baseline methods \textbf{MT} and \textbf{Bi-AFCG} use the old evaluation criterion, while other baseline methods \textbf{LSTM}, \textbf{S2VT}, \textbf{TA}, \textbf{H-RNN}, \textbf{DCE} and \textbf{DVC} use the new evaluation criterion. For fair comparison, we provide our results by using both criteria. From Table~\ref{table:learned}, our \textbf{GPaS} method achieves the highest scores in terms of all metrics when using different versions of criteria. It also outperforms some recent deep video captioning methods, such as \textbf{Bi-AFCG}, \textbf{MT}, and \textbf{DVC}.

We further compare the captioning performance on the testing set of ActivityNet Captions. For fair comparison, we compare our  \textbf{GPaS} method with \textbf{Bi-AFCG} \cite{wang2018bidirectional}, in which for both methods we use the same C3D feature without ensembling multiple models. The online test server only reports the Meteor scores.
As shown in Table \ref{table:test}, our proposed approach achieves a Meteor score of $6.41\%$,  which outperforms the baseline method \textbf{Bi-AFCG}.

As shown in Table \ref{table:Youcook II}, our proposed method using the output of the ‘flatten-673’ layer in Resnet-200 and the optical flow features also achieves better performance than the baseline method in 
\cite{zhou2018end} on the  Youcook II validation dataset, which demonstrates the effectiveness of our proposed method for handling the videos with longer duration. 

\begin{table}[ht]
\centering
{
\caption{Performance comparison on the ActivityNet Captions testing dataset.}\label{table:test}
\begin{tabular}{|c|c|}
\hline
Model  & M\\
\hline
Bi-AFCG \cite{wang2018bidirectional} & 4.99  \\
\hline
GPaS (ours)  & \textbf{6.41}  \\
\hline
\end{tabular}
}
\end{table}

\begin{table}[ht]
\begin{minipage}{1\linewidth}
\centering
\caption{Performance comparison on the YouCook II validation dataset.}
\label{table:Youcook II}
\begin{tabular}{|c|c|c|c|c|}
\hline
method & R & B@4 & M & C \\
\hline
MT \cite{zhou2018end} & - & 1.42 & 11.20 & - \\
\hline
GPaS (ours) & 27.98 & \textbf{1.64}  & \textbf{12.20} & 41.44 \\
\hline
\end{tabular}
\label{table:yc2}
\end{minipage}
\end{table}

\subsection{Ablation  Study}
In our graph-enhanced summarization module, we propose two types of graphes (the basic graph and the extension graph) and two types of GLI modules between GCN and LSTM (GCN inside LSTM and GCN outside of LSTM). Here we conduct ablation study to investigate the contributions of each component in our model. In addition, we also compare our method with some basic summarization approaches, as explained below. The results are reported in Table~\ref{table:ablation1}.

\begin{table}[!ht]
\centering
{
\caption{Ablation study on the ActivityNet Captions validation dataset by using the ground-truth proposals.}\label{table:ablation1}
\begin{tabular}{|c|c|c|c|c|}
\hline
Model  & R & B@4   & M & C \\
\hline
TA (our implementation) \cite{yao2015describing}   & 19.21 & 1.42  & 9.14  &29.60 \\
\hline
PM-ave   & 18.86 & 1.22    & 8.46 & 24.52  \\
\hline
PM-best   & 18.98 & 1.34   & 9.07 & 28.91  \\
\hline
PaS-basic-w/o TVJ   & 19.12 & 1.43  & 9.00 & 28.11 \\
\hline
PaS-basic   & 19.26 & 1.53   & 9.30  & 28.66\\
\hline
aGCN-out-LSTM-basic  & 19.45 & \textbf{1.64}  & 9.46 & 29.12\\
\hline
aGCN-out-LSTM-expanded  & 19.73 & 1.52  & 9.62 & 29.71\\
\hline
aGCN-in-LSTM-basic   & 19.72  & 1.62  & 9.48 & 30.01\\
\hline
 aGCN-in-LSTM-expanded   & 20.50 & 1.60  & 9.72 & \textbf{31.11} \\
\hline
GPaS (ours)   & \textbf{21.30} & 1.53  & \textbf{11.04} & 28.20 \\
\hline
\end{tabular}
}
\end{table}

\textbf{TA} (our implementation) is our implementation of the temporal-attention model~\cite{yao2015describing}, which generates the captions for the proposals without video segment partition. 
\textbf{PM-ave} directly utilizes the segment-level sentences output by our partition module without summarization, and the mean Meteor score averaged over the scores from all segment-level sentences is reported. Different from \textbf{PM-ave}, \textbf{PM-best} selects a representative sentence with the highest confidence score\footnote{The confidence score is calculated as  $\frac{1}{M} \sum\limits_{i=1}^M \log (p(\textbf{w}_i))$, where $p(\textbf{w}_i)$ represents the probability of each predicted word in our module and $M$ is the number of words in the generated sentences.} as the final description of each event proposal.
\textbf{PaS-basic} and \textbf{PaS-basic-w/o TVJ} use the basic summarization modules with or without TVJ module, respectively. 
\textbf{aGCN-out-LSTM-basic} and \textbf{aGCN-out-LSTM-expanded} use the GLI modules based on the basic and the expanded graph, respectively, where the node of aGCN is outside the LSTM cell.
\textbf{aGCN-in-LSTM-basic} and \textbf{aGCN-in-LSTM-expanded} use the GLI modules based on the basic and the expanded graph, respectively, where the node of aGCN is inside of the LSTM cell. 
\textbf{GPaS} is our final approach, which is based on the result from \textbf{aGCN-in-LSTM-expanded}, but we additionally use reinforcement learning to generate better sentences and employed the confidence score to select fewer number of sentences as in \cite{wang2018bidirectional}. 
 From Table ~\ref{table:ablation1}, we have the following observations.

\begin{figure}[t]
    \centering
    \includegraphics[scale = 0.7]{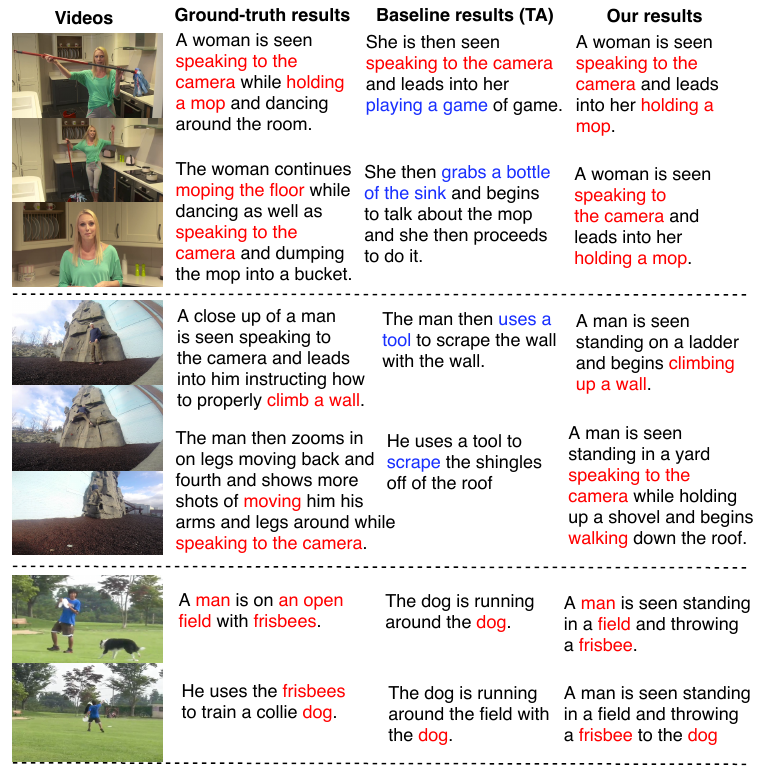}
    \caption{Generated sentences by using our method \textbf{aGCN-in-LSTM-expanded} and the baseline method \textbf{TA} \cite{yao2015describing} on the validation set of the ActivityNet Captions dataset. The correct semantic words are highlighted in red color, while the incorrect semantic words are highlighted in blue color.} 
    \label{visual1}
\end{figure}

First, with the same two-layer ``encoder-decoder" structure, the approaches \textbf{aGCN-out-LSTM-basic} and \textbf{aGCN-in-LSTM-basic} using the GCN-enhanced summarization module outperform \textbf{PaS-basic} using the basic summarization module. This result demonstrates it is effective to exploit word relationship for summarization.  Moreover, the approaches using the expanded graph for modeling the word relationship outperform those using the basic graph, when comparing the Meteor scores of \textbf{aGCN-in-LSTM-expanded} ($9.72\%$) with \textbf{aGCN-in-LSTM-basic} ($9.48\%$) and \textbf{aGCN-out-LSTM-expanded} ($9.62\%$) with \textbf{aGCN-out-LSTM-basic} ($9.46\%$), respectively.

Second, comparing \textbf{PaS-basic} and \textbf{PaS-basic-w/o TVJ}, we observe that fusion of visual features and semantic words can bring additional $0.3\%$ increase in terms of Meteor score, which shows it is beneficial to use the visual cues.

Third, our \textbf{GPaS} approach outperforms \textbf{TA} that does not utilize the PaS strategy. Our approach also wins \textbf{PM-ave} that does not conduct sentence summarization and \textbf{PM-best} that simply selects a representative segment-level sentence as the result. These results demonstrate the effectiveness of our \textbf{GPaS} framework and the importance of a powerful sentence summarization model for dense video captioning.

Finally, our \textbf{GPaS} method outperforms \textbf{aGCN-in-LSTM-expanded}, which demonstrates it is beneficial to apply reinforcement learning and the confidence score as in \cite{wang2018bidirectional}  for dense video captioning.

\begin{table}[!ht]
\centering
\caption{Performance of our method when using different values for the parameters $\lambda_{d}$, $L_{k}$ and $L_{m}$ on the ActivityNet Captions validation dataset.}
\begin{tabular}{|c|c|c|c|c|c|}
\hline
\multicolumn{2}{|c|}{Hyperparameter value}   & R & B@4  & M & C \\
\hline
&0  &19.43 &1.56 &  9.45 & 29.13 \\
\cline{2-6}
$\lambda_{d}$ &0.01 &19.89 &1.51 &  9.57 &29.89 \\
\cline{2-6}
&0.1 & 19.73 & 1.52 &  9.62  &29.71 \\
\cline{2-6}
&1  &19.71 &1.61 &   9.55 &29.86 \\
\hline
&15 &19.82 &1.57 &  9.54 &29.65 \\
\cline{2-6}
$L_{m} $ &20 & 19.73 & 1.52 &  9.62 &29.71\\
\cline{2-6}
&30 &19.66 &1.53 &  9.57 &29.98 \\
\hline
&20 &19.98 &1.58 &  9.52 &29.81 \\
\cline{2-6}
$L_{k}$  &25 & 19.73 & 1.52 &  9.62 &29.71 \\
\cline{2-6}
&30 &19.88 &1.60 &  9.57  &29.86\\
\hline
\end{tabular}
\label{table:hyper}
\end{table}

\color{black} In Table \ref{table:hyper}, we also evaluate the sensitivity of our proposed method when using different hyper-parameters. As shown in Table \ref{table:hyper}, the Meteor score  decreases without using the discriminative loss (i.e. $\lambda_{d} = 0$), which shows it is necessary to use the discriminative loss. Also, our proposed method is relatively stable when varying the hyper-parameters $L_m$, $L_k$, $\lambda_d$ in certain ranges, which demonstrates the robustness of our proposed method. \color{black}

\begin{figure}[t]
    \centering
    \includegraphics[scale = 0.7]{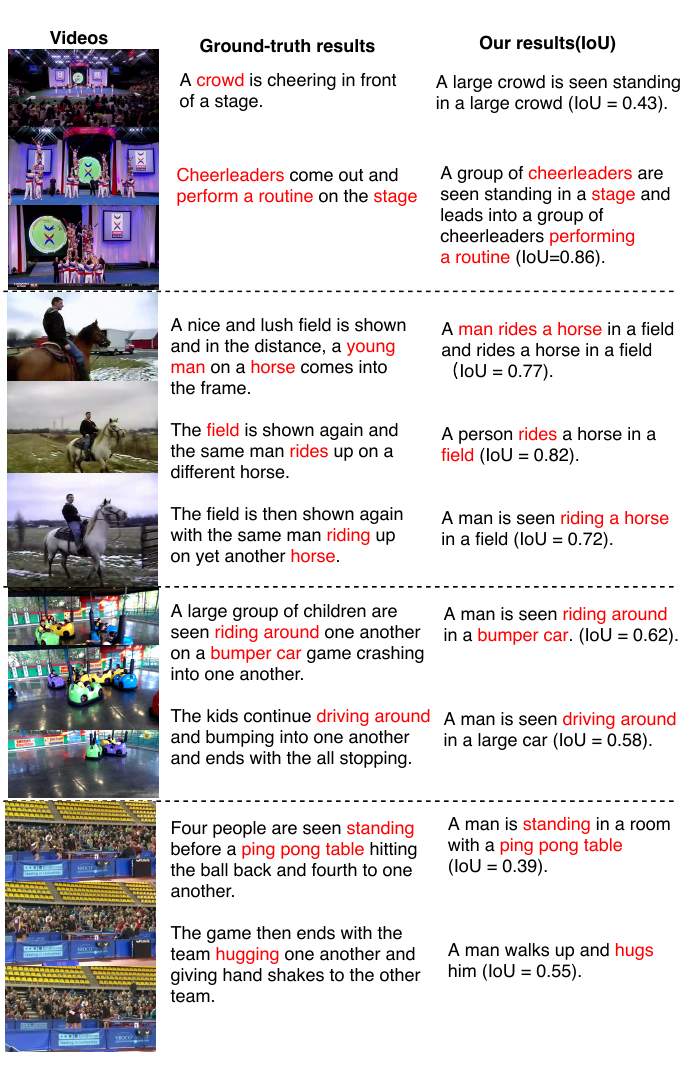}
    \caption{Examples of the  sentences generated by our \textbf{aGCN-in-LSTM-expanded} method when  using the automatically generated proposals. The IoU value with the ground-truth proposal is also reported after each sentence. The correct semantic words are highlighted in red color.} 
    \label{visual2}
\end{figure}

\begin{figure}[t]
    \centering
    \includegraphics[scale = 0.7]{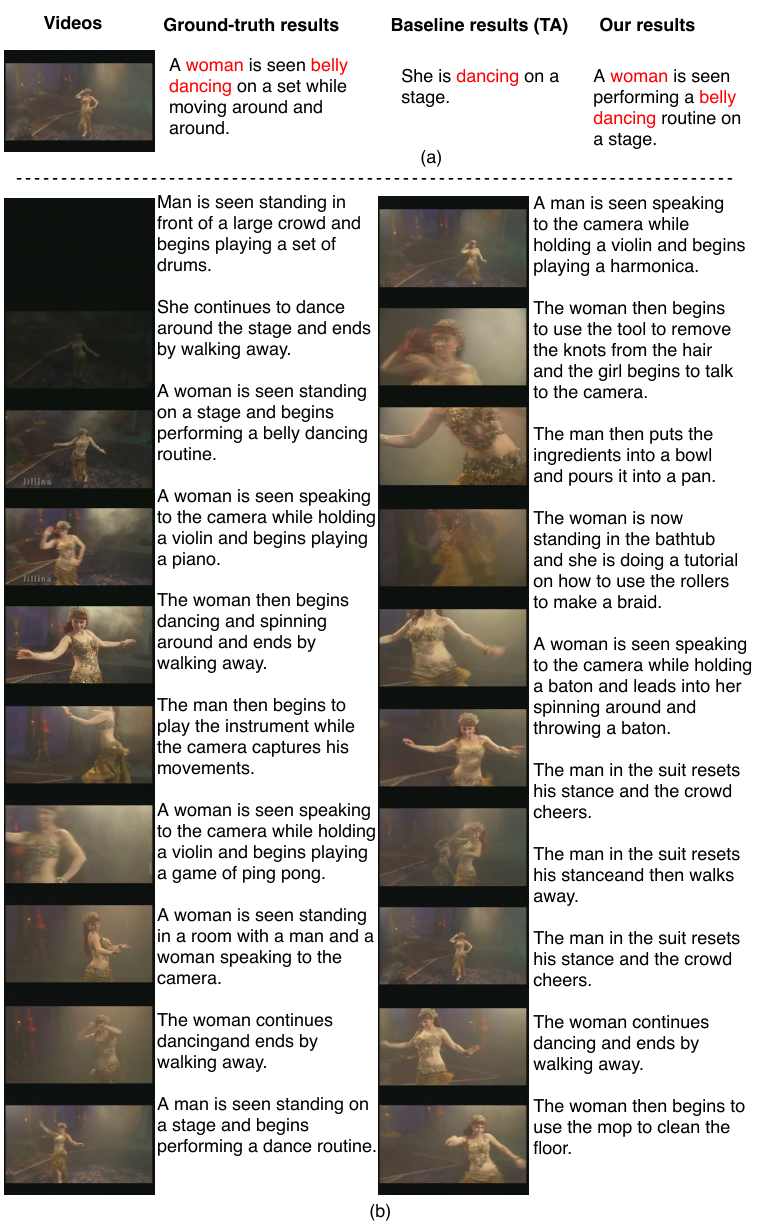}
    \caption{ (a) One sentence descriptions generated by our \textbf{aGCN-in-LSTM-expanded} method and the baseline method \textbf{TA}~\cite{yao2015describing}, respectively. The correct semantic words are highlighted in red color. (b) The corresponding $L_m (=20)$ segment-level sentences output by our partition module, which are also used as the textual input of our GCN-enhanced summarization module.}
    \label{visual3}
\end{figure}

\subsection{Qualitative Analysis}

\begin{figure}[t]
    \centering
    \includegraphics[scale = 0.7]{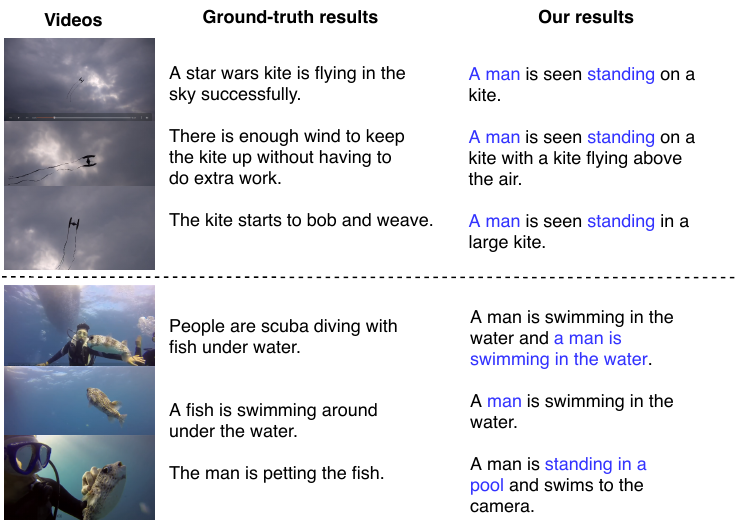}
    \caption{Some failure cases of our proposed method.} 
    \label{neg}
\end{figure}

Some visual examples of our dense video captioning results are shown in Fig. \ref{visual1}, Fig. \ref{visual2}, Fig. \ref{visual3} and Fig.\ref{neg}.

In Fig. \ref{visual1}, we compare the sentences generated by our  \textbf{aGCN-in-LSTM-expanded} method and the baseline method \textbf{TA} \cite{yao2015describing} when using the ground-truth proposal. It can be observed that our proposed method can better describe the video content when compared with the baseline method.  Taking the first video in Fig. \ref{visual1} as an example, our method correctly generates ``speaking to the camera" and ``holding a mop" while the baseline wrongly generates ``playing a game" and ``grabs a bottle of 
the sink".

In Fig. \ref{visual2}, we show some examples of the sentences generated by using our \textbf{aGCN-in-LSTM-expanded} method when using the automatically generated proposals. The IoU value with the ground-truth proposal is also reported. From Fig. \ref{visual2}, it shows that our proposed method generates reasonable sentences even with automatically generated proposals.

In Fig. \ref{visual3}, we provide an example of the sentences generated by our \textbf{aGCN-in-LSTM-expanded} method and the \textbf{TA} method (Fig.~\ref{visual3} (a)), as well as the corresponding $L_m(=20)$ segment-level sentences output by our partition module, which are also used as the textual input of our GCN-enhanced summarization module (Fig.~\ref{visual3} (b)). Although some of the $L_m$ segment-level sentences can not produce correct descriptions (e.g. ``belly dancing" is wrongly described as ``playing piano" and ``cleaning the floor"),  a better sentence description could still be obtained by using our \textbf{aGCN-in-LSTM-expanded} method, when compared with the baseline method \textbf{TA}.

We also provide some failure cases in Fig.\ref{neg}. In the first video, a crosslet on the kite is wrongly recognized as a person, leading to the wrongly generated caption ``a man is seen standing on a kite''. This issue could be addressed by using extra reliable object detection results. In the second video, the statement of  ``a man is swimming in the water'' wrongly appears twice in one sentence, and also produces redundant information, which could be solved by introducing a penalty term to discourage the repeated statements. These issues will be studied in our future work

\color{black}

\section{Conclusion}
With the proposed GPaS framework, we utilize the visual and semantic features at the sub-proposal level to generate the one-sentence description for the whole event proposal. Particularly, an GCN-enhanced summarization module is proposed for sentence summarization. It utilizes the word relationship to refine the hidden representations of the generated sentence. Specifically, we treat the semantic words as the nodes in a GCN graph, and learn their interactions via the tightly coupled GCN and LSTM networks. From our experiments, we observe that better message transfer in the hidden representations/cell states of different LSTM cells can be achieved by connecting GCN to each LSTM cell. In addition, the gain obtained by using expanded graph shows that we can generate better sentences by exploiting different levels of semantic meanings in the dense video captioning task.

% or


%\appendices

%\section{}

%\section*{Acknowledgment}


\ifCLASSOPTIONcaptionsoff
  \newpage
\fi



%\bibliographystyle{IEEEtran}

{\small
\bibliographystyle{IEEEtran}
\bibliography{main}
}
\vspace{-15mm}
\begin{IEEEbiography}[{\includegraphics[width=1in,height=1.25in,clip,keepaspectratio]{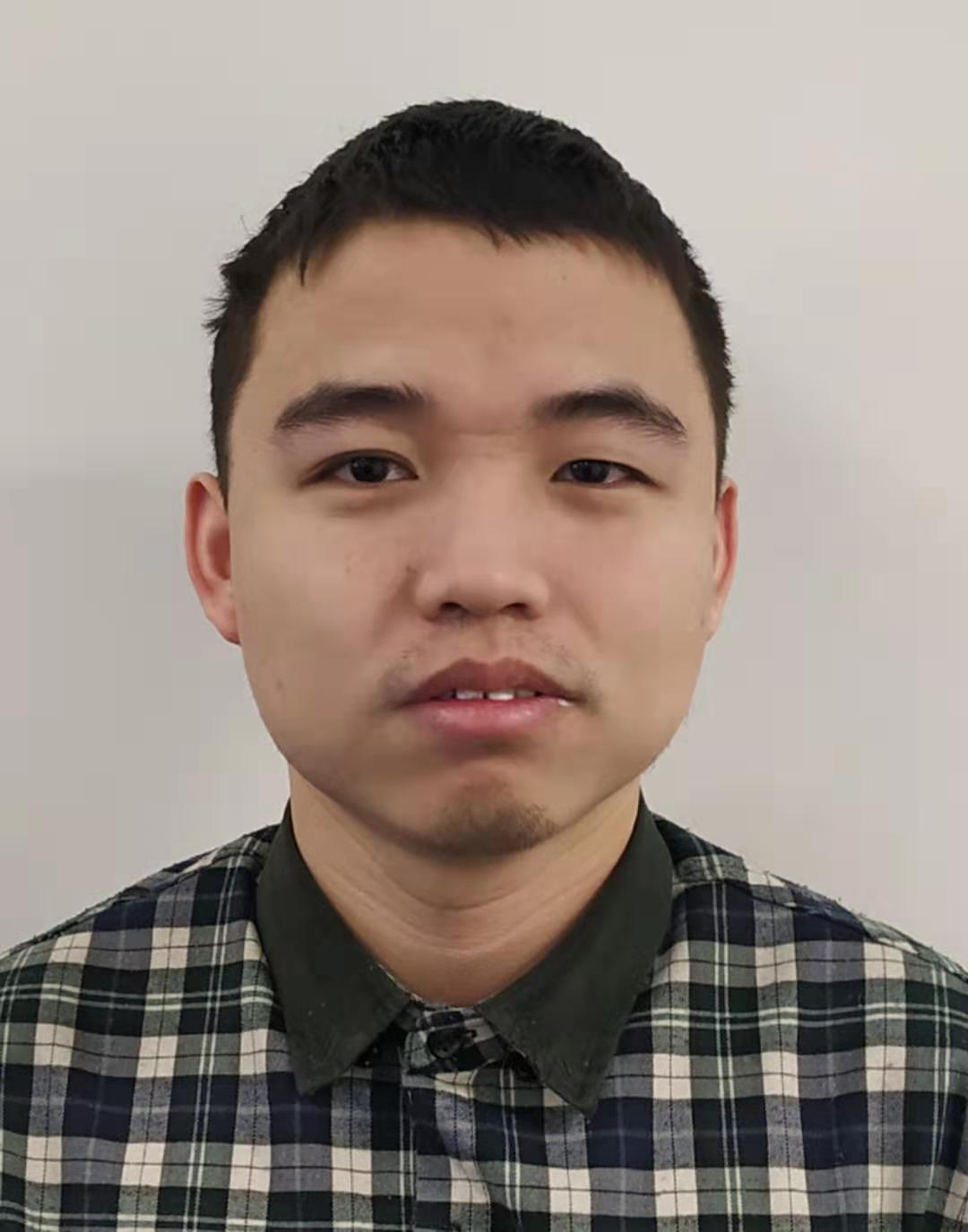}}]{Zhiwang Zhang}
received BE degree in school of Electrical and Information Engineering from the University of Sydney in 2017, where he is currently pursuing the PhD degree. His research interests include deep learning and its applications on computer vision and natural language processing. 
\end{IEEEbiography}

\vspace{-15mm}
\begin{IEEEbiography}[{\includegraphics[width=1in,height=1.25in,clip,keepaspectratio]{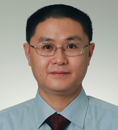}}]{Dong Xu} (F’18) 
 received the BE and PhD degrees from University of Science and Technology of China, in 2001 and 2005, respectively. While pursuing the PhD degree, he was an intern with Microsoft Research Asia, Beijing, China, and a research assistant with the Chinese University of Hong Kong, Shatin, Hong Kong, for more than two years. He was a post-doctoral research scientist with Columbia University, New York, NY, for one year. He worked as a faculty member with Nanyang Technological University, Singapore. Currently, he is a professor and chair in Computer Engineering with the School of Electrical and Information Engineering, the University of Sydney, Australia. His current research interests include computer vision, statistical learning, and multimedia content analysis. He was the co-author of a paper that won the Best Student Paper award in the IEEE Conferenceon Computer Vision and Pattern Recognition (CVPR) in 2010, and a paper that won the Prize Paper award in IEEE Transactions on Multimedia (T-MM) in 2014. He is a fellow of the IEEE.
\end{IEEEbiography}
\vspace{-15mm}
\begin{IEEEbiography}[{\includegraphics[width=1in,height=1.25in,clip,keepaspectratio]{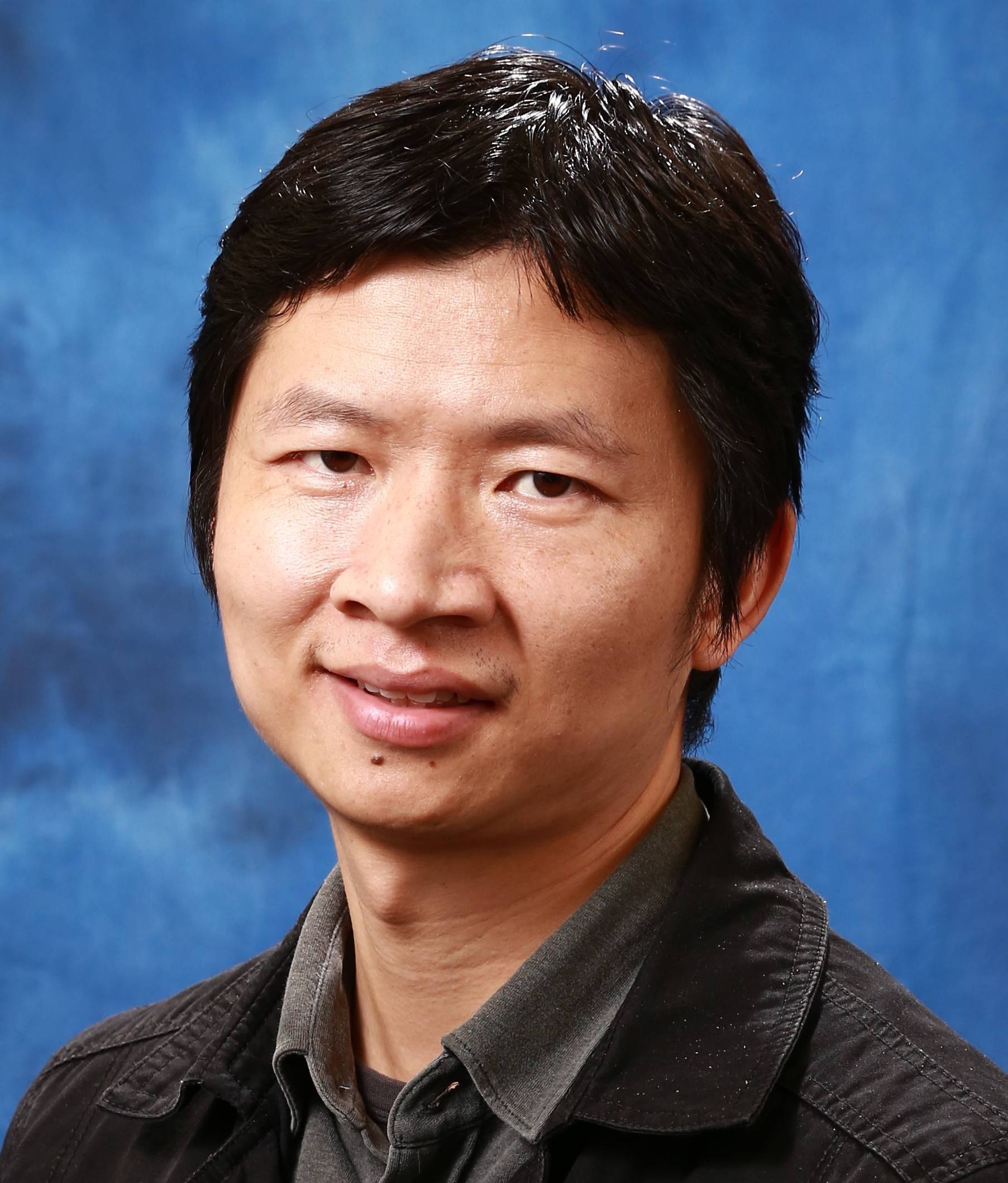}}]{Wanli Ouyang} (SM’16)
received the PhD degree in the Department of Electronic Engineering, Chinese University of Hong Kong. Since 2017, he is a senior lecturer with the University of Sydney. His research interests include image processing, computer vision, and pattern recognition. He is a senior member of the IEEE.
\end{IEEEbiography}
\vspace{-15mm}
\begin{IEEEbiography}[{\includegraphics[width=1in,height=1.25in,clip,keepaspectratio]{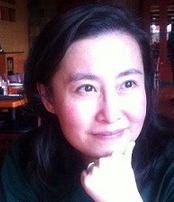}}]{Luping   Zhou} (SM’13) received the Ph.D. degree from Australian  National University. She is currently a Senior Lecturer  with the School of Electrical and Information Engineering, The University of Sydney, Australia. Her research interests include machine learning, computer vision, and medical image  analysis. She was a recipient of the Australian Research Council DECRA award (Discovery Early Career  Researcher Award) in 2015.
\end{IEEEbiography}

%\end{IEEEbiography}


%\newpage



%\vfill


\end{document}